# Comparative Evaluation of Traditional and Deep Learning Feature Matching Algorithms using Chandrayaan-2 Lunar Data

**Ridhi Makharia[1], Jai G Singla[2], Amitabh[3], Nitant Dube[4], Harish Sharma[5]**

[1,5] School of Computer Science Engineering, Department of AIML, Manipal University Jaipur, Rajasthan, India

[2,3,4] Signal and Image Processing Area, Space Applications Centre ISRO, Ahmedabad, Gujarat, India

**Abstract**

Accurate image registration is critical for lunar exploration, enabling surface mapping, resource localization, and mission planning. Aligning data from diverse lunar sensors- optical (e.g., Orbital High Resolution Camera, Narrow and Wide Angle Cameras), hyperspectral (Imaging Infrared Spectrometer), and radar (e.g., Dual-Frequency Synthetic Aperture Radar, Selene/Kaguya mission)- is challenging due to differences in resolution, illumination, and sensor distortion. We evaluate five feature matching algorithms: SIFT, ASIFT, AKAZE, RIFT2, and SuperGlue (a deep learning-based matcher), using cross-modality image pairs from equatorial and polar regions. A preprocessing pipeline is proposed, including georeferencing, resolution alignment, intensity normalization, and enhancements like adaptive histogram equalization, principal component analysis, and shadow correction. SuperGlue consistently yields the lowest root mean square error and fastest runtimes. Classical methods such as SIFT and AKAZE perform well near the equator but degrade under polar lighting. The results highlight the importance of preprocessing and learning-based approaches for robust lunar image registration across diverse conditions.

## 1. INTRODUCTION

Image registration is a critical preprocessing step in lunar remote sensing, ensuring spatial alignment between multiple images of the same region. This enables accurate comparisons, data fusion from different sensors, and change detection over time. With the growing volume and diversity of lunar datasets from missions like Chandrayaan-2 and LRO, robust registration methods are essential for unlocking their full scientific value. [1, 2] Inaccurate registration can introduce errors that compromise downstream analyses and mission planning. As future missions produce increasingly complex data, the demand for precise registration will continue to grow.

High-accuracy registration underpins key lunar science tasks such as map and mosaic generation, landing site assessment, and hazard analysis. Aligning imagery from sensors like OHRC, LRO NAC/WAC, DFSAR, and SELENE allows for detailed surface monitoring, spectral-spatial data integration, and reliable geological mapping.

However, lunar image registration faces unique challenges. Geometric differences arise from varying sensor angles, orbits, and calibrations, while radiometric disparities result from changing illumination, sensor properties, and the Moon's surface characteristics, particularly in polar regions with low solar angles and permanent shadows.[2] The Moon's harsh environment—absence of atmosphere, extreme temperature variations, and cratered terrain—further complicates feature detection and matching.

The core goal of registration is to ensure that corresponding pixels across images refer to the same physical location. This involves geometric correction—transforming images through translation, rotation, scaling, affine, or polynomial models—and radiometric correction to adjust pixel intensities for consistent brightness and contrast.

## 2. LITERATURE REVIEW

This literature survey focuses on understanding feature matching algorithms—SIFT (Scale-Invariant Feature Transform), ASIFT (Affine-SIFT), AKAZE (Accelerated-KAZE), RIFT2 (Radiation Insensitive Feature Transform 2), and SuperGlue—and reviews previous work on image registration of lunar data using these methods. It also aims to highlight the limitations of these algorithms in order to underscore the relevance of the proposed solution.

### 2.1. Known Limitations of Image Registration Algorithms

SIFT is a traditional feature matching algorithm known for its robustness to changes in scale and rotation. However, its multi-stage process and large descriptor size (128 dimensions) lead to relatively high computational overhead.[5] The multiple stages involved and the high dimensionality of the descriptor vectors make SIFT relatively slower compared to some other feature detection and description methods.[3] This can be a limitation in applications requiring real-time processing or when dealing with very large images.[3] Additionally, the performance of SIFT may degrade in image regions that lack distinct features or when there are significant changes in the viewpoint between the images being registered.[6]

ASIFT extends SIFT by simulating various camera angles to achieve full affine invariance. This dramatically improves its robustness to perspective changes. However, ASIFT is significantly more computationally expensive than SIFT because it involves performing multiple simulations of the image and applying the SIFT algorithm to each of these simulated views.[7] The computational complexity of ASIFT can be approximately twice that of SIFT, or even substantially higher, depending on the number of simulated viewpoints.[8]

RIFT2 is an optimization of the RIFT algorithm that is designed for matching images taken under different lighting conditions.[3] It uses radiation-insensitive descriptors, which make it ideal for registering cross-sensor images.[9] It is significantly more efficient than the original RIFT due to rotation-invariance mechanism.[10] Still, it may struggle with severe geometric distortions or in areas that lack distinct image features.[10]

AKAZE operates within a nonlinear scale space and uses a lightweight binary descriptor.[11] It performs faster than both SIFT and its predecessor, KAZE, making it an ideal choice for real-time applications.[13] However, its not as effective as SIFT in case of extreme scale and geometric distortions.[13]

SuperGlue introduces a deep learning-based approach that uses a neural network to match image features by learning spatial relationships between them. It performs well in real-time when a GPU is available and can outperform traditional methods in feature matching accuracy.[14] However, SuperGlue's efficiency is highly dependent on the quality and quantity of initial features (e.g., from SuperPoint) and requires domain-specific training for optimal results. In resource-limited environments or uncommon domains, these dependencies can become bottlenecks.[16]

**2.2. Past Work on Lunar Image Registration Using Specific Datasets and Algorithms**

Researchers have studied the registration of high-resolution panchromatic images from Chandrayaan-2's OHRC and LRO's NAC to improve lunar mapping. Algorithms like SIFT and ASIFT have achieved sub-pixel accuracy (0.2–2.5 pixels), but challenges arise from shadowing and subtle intensity differences on the lunar surface. These have been mitigated through preprocessing techniques that enhance edge and corner detection.[17]

Several studies have focused on the co-registration of hyperspectral data with the globally geo-referenced LROC WAC mosaic.[18] Challenges here include differences in spectral range and spatial resolution between sensors. Methods such as manually selecting homologous points to compute an initial affine transformation, or applying RANSAC-based models followed by polynomial refinement, have proven effective. [18, 19] These techniques have achieved registration accuracy comparable to manual alignment.

The registration of DFSAR data with SELENE or other optical/altimetry data often involves indirect comparisons or specialised analysis of radar scattering properties.[20] A consistent trend is the challenge posed by significant differences in sensor characteristics and resolution, requiring tailored approaches for each dataset combination.

## 2.3. Comparative Performance Analysis of Image Registration Algorithms on Lunar Datasets

SIFT and ASIFT have shown strong performance for high-resolution optical image alignment, while affine and orientation-based methods have been effective for aligning hyperspectral and multispectral datasets. For radar-optical comparisons, specialised or indirect methods are often required. [16, 18, 20]

Comparative studies, such as those using the SyncVision framework, have evaluated algorithms like SIFT, ORB, and custom methods like IntFeat across scales. SIFT offered high accuracy but required more processing time; ORB was faster but less precise. AKAZE delivered a good compromise between speed and accuracy. [3]

Emerging deep learning methods—especially GNN-based models like SuperGlue—show promise for handling complex lunar imagery. However, their reliance on extensive, well-labelled training data, GPU resources, and potential for overfitting or reduced interpretability remains a challenge. [16, 23]

## 2.4. Gaps in Current Research and Contribution of the User's Paper

Despite ongoing research, few studies have compared a full suite of traditional and deep learning-based algorithms, such as SIFT, ASIFT, RIFT2, AKAZE, and SuperGlue, across specific lunar datasets like OHRC–LROC NAC, IIRS–LROC WAC, and DFSAR–SELENE, particularly in polar regions where terrain and illumination are more extreme.

This work addresses that gap through a systematic, multi-algorithm comparison across key dataset pairs in both equatorial and polar zones. It evaluates the trade-offs in accuracy, robustness to distortions, and computational cost, offering valuable guidance for choosing the most suitable registration strategy for lunar remote sensing applications.

This comparative analysis will advance the field of lunar image registration by providing a much-needed benchmark for these widely used algorithms on critical lunar datasets, ultimately contributing to more accurate and efficient utilisation of lunar remote sensing data for scientific discovery and exploration.

## 3. LUNAR DATASETS

This research focuses on registering heterogeneous data from different instruments for a synergised environment. The datasets selected span across panchromatic, spectral and radar domains of multiple spatial resolutions. Table 1 represents the detailed overview of the datasets used.

### 3.1. OHRC and LROC NAC Registration

The Orbiter High Resolution Camera (OHRC) is a panchromatic imaging system that works within the visible spectrum. It can achieve a Ground Sampling Distance (GSD) of 0.3 m. To accurately register the data, we align it with NASA's LROC Narrow Angle Cameras (NAC). It has 2 high-resolution cameras that can capture the lunar surface at approximately 0.5 meters to 1 meter.

### 3.2. IIRS and LROC WAC Registration

The Imaging Infrared Spectrometer (IIRS) is a hyperspectral imaging system that has a lower spatial resolution—around 80 meters. It is registered with the LROC Wide Angle Camera (WAC). The WAC offers a global coverage of the lunar surface at a spatial resolution of 100 meters per pixel.

### 3.3. DFSAR and SELENE Registration

The Dual-Frequency Synthetic Aperture Radar (DFSAR) is a microwave imaging system that has a high spatial resolution, ranging from 2 meters to 75 meters. It is registered with Japan's SELENE (Kaguya) mission. The SELENE dataset includes images from the Terrain Camera (TC) and the Multiband Imager (MI). The TC gives a spatial resolution of 10 meters, while the MI provides spectral imaging in visible and near-infrared bands, with resolutions ranging from 20 to 62 meters.

## 4. METHODOLOGY

This methodology outlines a strong approach for registering various lunar datasets through specialized preprocessing techniques, feature extraction, and image alignment. Our strategy tackles the specific challenges of lunar image registration by using tailored enhancement methods for different sensor types to boost feature detection in the tricky lunar environment.

Initial attempts to use feature detection algorithms like SIFT, ASIFT, and AKAZE didn't yield the desired results, which is mainly due to the unique challenges presented by lunar imagery: there are no atmospheric effects, drastic variations in illumination, and the generally uniform nature of the lunar surface. The key points generated were often incorrect, leading to faulty matches. This meant we needed specialized preprocessing techniques to enhance distinct features—like craters and edges—before we could effectively extract features. Figure 1 summarizes the overall methodology used for the image registration process.

|  | Equatorial Region | | Polar Region | |
|---|---|---|---|---|
|  | **OHRC** | **NAC** | **OHRC** | **NAC** |
| **ID** | ch2_ohr_ncp_20210401T2357376656_d_img_d18 | M1350459544RE | ch2_ohr_ncp_20200824t0806596861_d_img_d18 | M165491149RE |
| **Resolution** | 0.2648 | 1.1179 | 0.258557 | 0.88779 |
| **Height** | 90148 | 78252 | 90148 | 29696 |
| **Width** | 12000 | 8669 | 12000 | 2532 |
|  | **IIRS** | **WAC** | **IIRS** | **WAC** |
| **ID** | ch2_iir_nci_20220221T1109265965_d_img_d18 | WAC Global Morphologic Map | ch2_iir_nci_20230620T0234388613_d_img_d32 | WAC Global Morphologic Map |
| **Resolution** | 71.24 | 100 | 89.72 | 100 |
| **Height** | 2812 | 2812 | 8896 | 8896 |
| **Width** | 250 | 250 | 250 | 250 |
|  | **DFSAR** | **SELENE** | **DFSAR** | **SELENE** |
| **ID** | ch2_sar_ncxl_20200818t081301688_d_sri_xx_fp_xx_d18 | TCO_MAPs02_N21E312N18E315SC | ch2_sar_ncxl_20230906t025856273_d_cp_d32 | TCO_MAPs02_S69E030S72E033SC |
| **Resolution** | 20.52341, 19.986164 | 8.4232 | 5.120307, 1.998616 | 8.4232 |
| **Height** | 6783 | 10800 | 1804 | 10800 |
| **Width** | 734 | 10800 | 373 | 10800 |

*Table 1.* Product ID, resolution (meters per pixel), height (pixels) and width (pixels) of the datasets. Resolution specifications of DFSAR datasets, showing along-track and across-track resolutions in meters per pixel.

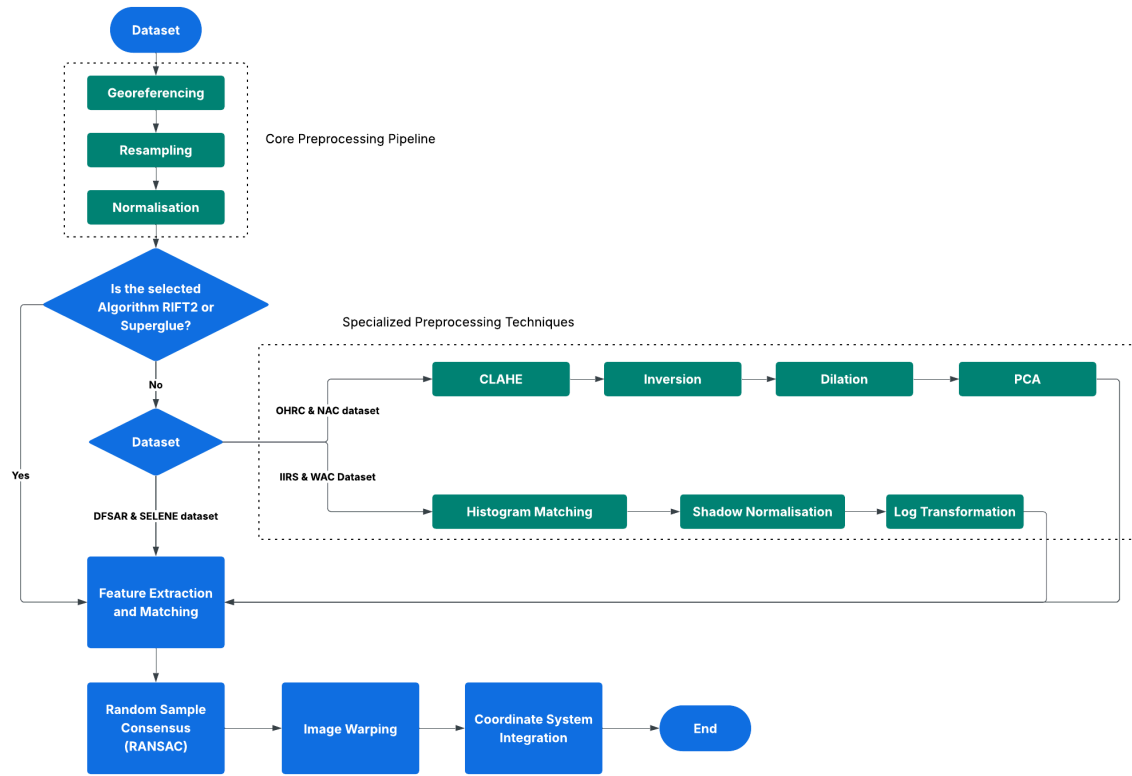

*Figure 1.* Methodology Flowchart

## 4.1. Core Preprocessing Pipeline

### 4.1.1. Georeferencing

Georeferencing was the first step taken to ensure consistent spatial referencing across the different lunar datasets. This involved aligning satellite imagery with actual lunar geographic coordinates.

The OHRC datasets used a Selenographic projection, whereas the LROC NAC had a projection of Equirectangular Moon. It was necessary to align the geographic projections; therefore, the OHRC datasets' spatial coordinates were georeferenced to match LROC NAC. This process was applied to other datasets, matching IIRS to WAC and DFSAR to SELENE.

### 4.1.2. Resolution Resampling

The datasets had significant resolution disparities that required normalisation before registration:

- OHRC (~30 cm resolution) was resampled to match NAC's resolution (0.5-2.0 m/pixel)
- IIRS (~80 m resolution) was resampled to align with WAC's 100 m/pixel resolution
- DFSAR (2-75 m slant resolution) was resampled to match SELENE's resolution (~9 m/pixel)

Resampling ensured that features appeared at comparable scales in both images, facilitating more accurate keypoint matching. This process involved interpolation to create new pixel values in the target resolution, allowing for direct pixel-to-pixel comparisons between image pairs.

### 4.1.3. Intensity Normalisation

All datasets were normalised to an 8-bit range (0-255) to ensure standardised intensity values across different sensor types. This normalization eliminated sensor-specific variations in radiometric scales, creating a consistent intensity range that established a level comparison basis for feature extraction algorithms.

## 4.2. Specialised Preprocessing Techniques

A. For OHRC and NAC Registration

### 4.2.1. Contrast Limited Adaptive Histogram Equalisation (CLAHE)

CLAHE was applied to balance intensity distributions across the lunar images while limiting noise amplification. Unlike traditional histogram equalisation that applies global contrast enhancement, CLAHE operates on small regions (tiles), enhancing local contrast while preventing over-amplification of noise in homogeneous regions.

For lunar imagery with extreme illumination variations, CLAHE proved particularly effective as it:

- Enhanced visibility of features in both shadowed and bright regions
- Limited contrast enhancement in relatively homogeneous regions to reduce noise amplification
- Preserved the edges of craters and other morphological features critical for matching

The implementation involved dividing images into tiles, calculating histograms for each tile, limiting the height of each histogram to reduce noise amplification, and then applying bilinear interpolation to eliminate artificial boundaries.

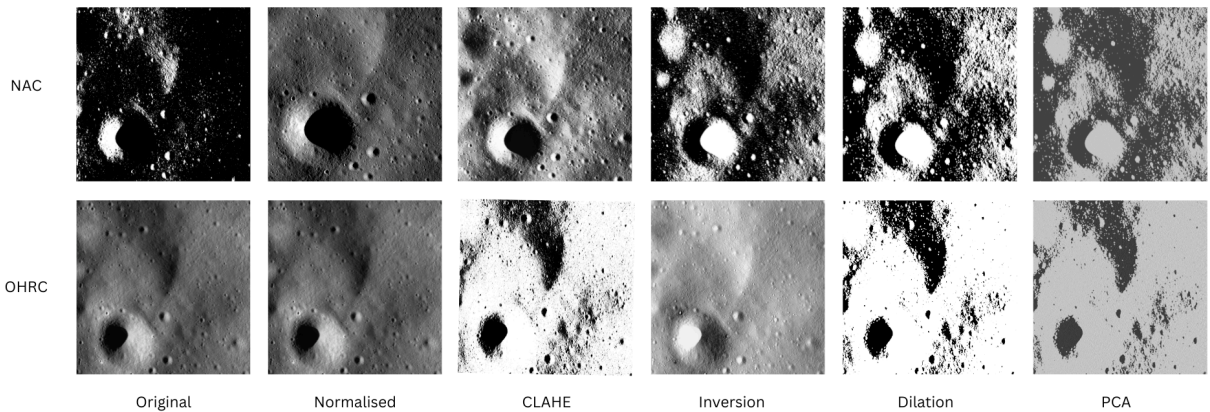

*Figure 2.* OHRC and NAC preprocessing steps visualisation

*4.2.2. Image Inversion*

The inversion process transformed pixel values to their complement (255 - pixel value), effectively highlighting features that were originally obscured. This technique enhanced the contrast between craters and the surrounding terrain, making rim structures more detectable for feature extraction algorithms.

*4.2.3. Morphological Dilation*

Dilation was applied to enhance the visibility of crater rims and elevated structures by expanding their boundaries. Using a defined structuring element, this morphological operation strengthened edge features, making them more prominent for feature detection. The enhanced edges provided stronger keypoints, improving the accuracy of subsequent feature matching.

*4.2.4. Principal Component Analysis (PCA)*

PCA was employed to reduce dimensionality while preserving the most significant variance in the data. This technique identified patterns in the lunar surface data and transformed the original variables into a new set of variables (principal components) that highlighted the most distinctive features while reducing noise. This preprocessing step was particularly valuable for highlighting patterns in the complex lunar terrain that might otherwise be difficult to detect using original image data.

  B.  *For IIRS and WAC Registration*

From the multi-band dataset, a single, visually clear band was selected as the reference for image registration. Once the transformation parameters were computed using this band, they were applied uniformly to align all other bands accordingly.

*4.2.1. Histogram Matching*

Histogram matching transformed the statistical distribution of pixel intensities in the IIRS imagery to match the histogram of the corresponding WAC imagery. This technique ensured radiometric consistency between the hyperspectral IIRS data and the multispectral WAC data, addressing differences in sensor calibration and illumination conditions.

By making the intensity distributions comparable, histogram matching facilitated more accurate feature detection across these different sensor types.

*4.2.2. Shadow Normalisation*

Shadow normalisation reduced the effects of lunar shadowing, which is particularly pronounced due to the lack of atmospheric light scattering on the Moon. This technique adjusted the intensity values in shadowed regions to make features more visible while preserving overall scene contrast.

The importance of this step cannot be overstated for lunar imagery, as the extreme shadow conditions can completely obscure surface features that are vital for registration. By normalizing shadows, we revealed features that would otherwise be lost in dark regions, significantly improving the capacity for feature matching.

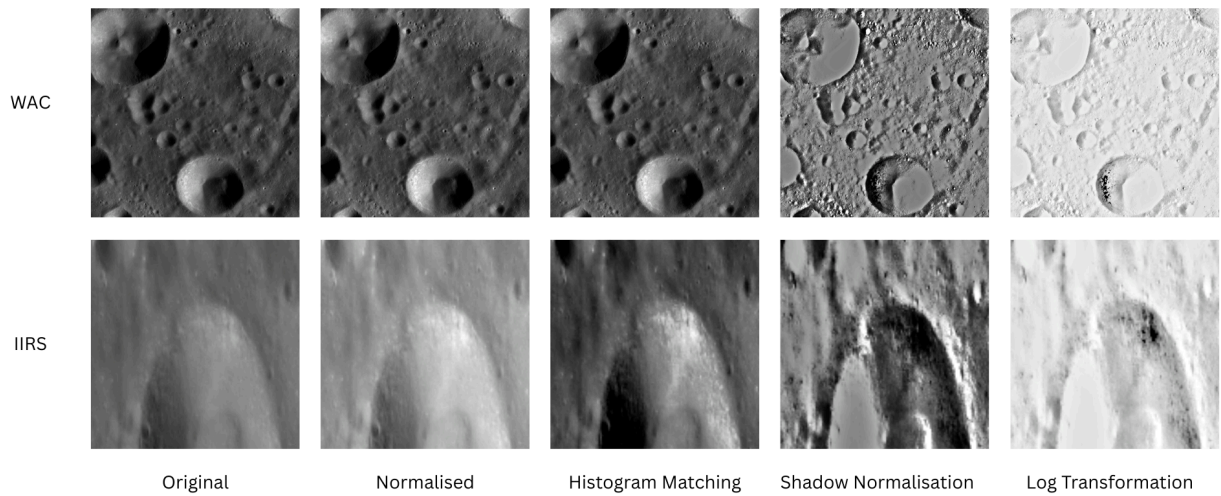

*Figure 3. IIRS and WAC preprocessing steps visualisation*

### 4.2.3. Log Transformation

Log transformation was applied to enhance low-intensity details while compressing higher-intensity values.

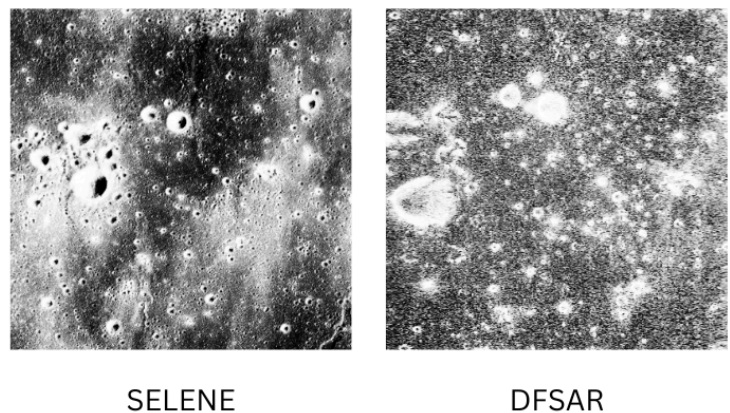

*Figure 4. SELENE and DFSAR sample dataset*

## 4.3. Feature Extraction and Matching

After applying the preprocessing techniques, feature extraction was performed using multiple algorithms to identify distinctive keypoints in the images:

### 4.3.1. Scale-Invariant Feature Transform (SIFT)

SIFT relies on constructing a scale-space using the Difference of Gaussians (DoG) to detect keypoints that remain stable under scale and orientation changes. It encodes local image structure using a 128-dimensional descriptor built from gradient orientations around each keypoint. [4]

### 4.3.2. Affine-SIFT (ASIFT)

ASIFT extends the concept of SIFT by artificially simulating camera tilt and rotation across a dense set of affine transformations. It applies SIFT on each simulated view, ensuring the detection of features that would otherwise be lost due to viewpoint changes.[8]

### 4.3.3. Radiation-insensitive Feature Transform (RIFT2)

RIFT2 uses phase congruency, a method based on local energy models rather than intensity gradients, to detect features that remain stable across different imaging modalities. Its descriptors are constructed using a maximum index map (MIM), which is more resilient to non-linear radiometric variations compared to traditional approaches. [10]

### 4.3.4. Accelerated-KAZE (AKAZE)

AKAZE operates in a nonlinear scale space created through Perona-Malik diffusion, enabling better edge preservation and noise suppression during detection. It uses a Modified-Local Difference Binary (M-LDB) descriptor that is compact and suitable for efficient matching. [11, 12]

### 4.3.5. SuperGlue

SuperGlue is built on deep learning principles. It employs graph neural networks to model spatial relationships between features and solves an optimal transport problem to identify the best matches between keypoints. This allows it to learn and generalise geometric priors from training data, which can lead to superior performance in complex scenes, especially when paired with a high-quality detector like SuperPoint.[14, 15]

## 4.4. Random Sample Consensus (RANSAC)

RANSAC was employed to ensure robust feature matching by eliminating incorrect correspondences (outliers) between the preprocessed images. The algorithm works iteratively by randomly selecting a subset of matched features and computing a homography transformation model based on those matches. It then evaluates the model by determining the number of inliers—feature matches that are consistent with the estimated transformation. This process is repeated multiple times, and the model yielding the highest number of inliers is selected as the most reliable, ensuring accurate geometric alignment between the image pairs.

RANSAC was crucial for lunar image registration as it effectively handled the high proportion of outliers that typically occur due to the repetitive nature of lunar features, illumination differences, and sensor variations.

## 4.5. Image Warping

The homography matrix calculated from the RANSAC-filtered feature matches was used to warp the source images (OHRC, IIRS, DFSAR) to align with their respective reference images (NAC, WAC, SELENE), as shown in Figure 5 and Figure 6. This 3×3 matrix transformation accounted for geometric differences between the image pairs, including translation, rotation, scale, and perspective distortions.

## 4.6. Coordinate System Integration

The final step in the registration process involved integrating the coordinate reference systems of the warped and reference images:

- The top-left coordinates of the reference image (NAC, WAC, SELENE) was determined from its metadata file.
- The coordinate system of the warped image (OHRC, IIRS, DFSAR) was adjusted to align with this reference point, and it was then converted to its original coordinate reference system, which was before the georeferencing step.
- This alignment integrated the spatial information of both images, which resulted in a coherent and accurately aligned lunar surface composite

This coordinate integration was essential for creating scientifically valid registered products that preserved the spatial accuracy of the original data while enabling direct comparison and analysis across different sensor types.

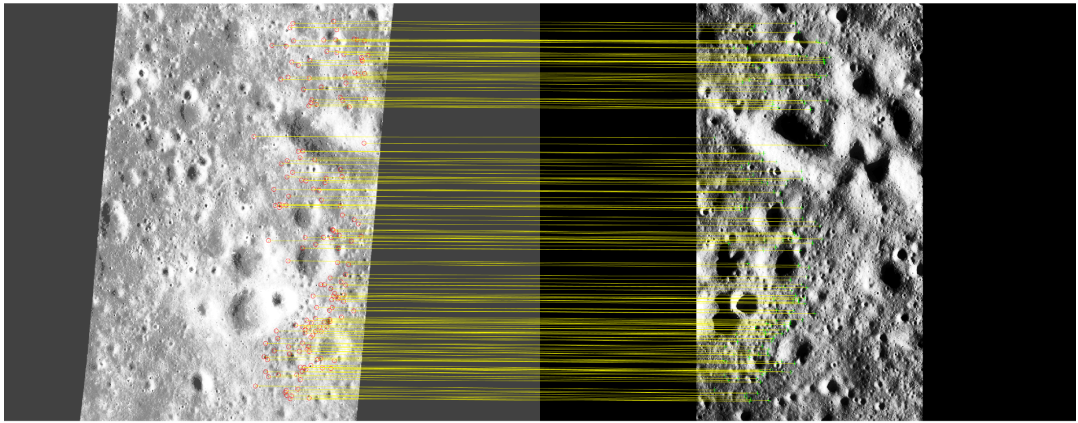

*Figure 5.* Result of applying the RIFT2 algorithm for feature extraction and matching (Step 4.3), followed by filtering the outliers using the RANSAC algorithm (Step 4.4) on OHRC and LRO's NAC. The yellow lines represent matched keypoints between the two images, with green and red points indicating the detected features in the respective images.

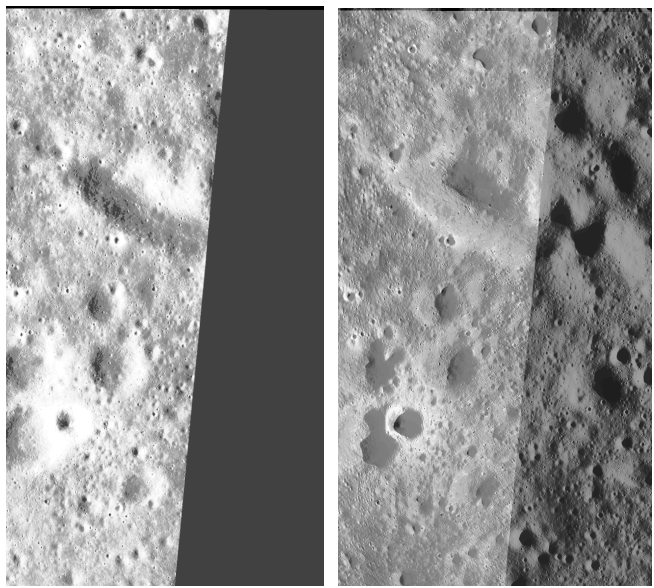

***Figure 6.*** *The left image shows the OHRC image after applying a perspective transformation and additional padding, enabling accurate geometric alignment with the NAC reference frame. The right image displays the stitched result, where the transformed OHRC image is overlaid onto the LROC NAC base. The alignment across shared features like craters and ridges demonstrates successful registration based on matched keypoints.*

**4.7. Evaluation Metrics**

*4.7.1. Root Mean Square Error (RMSE)*

$$\text{RMSE} = \sqrt{\frac{\sum_{i=1}^{N}(x_i - \hat{x}_i)^2}{N}}$$

$\text{RMSE}$ = root-mean-square error
$i$ = variable i
$N$ = number of non-missing data points
$x_i$ = actual observations time series
$\hat{x}_i$ = estimated time series

(1)

RMSE was used to quantitatively assess registration accuracy by measuring the average magnitude of errors between corresponding points in the reference and registered images. The process involved:

- Identifying matching control points in both the reference image and the transformed warped image
- Applying the derived perspective transformation to align the control points
- Calculating the Euclidean distance between each pair of corresponding points
- Computing the square root of the average of these squared distances

The RMSE was calculated based on Formula 1. Lower RMSE values indicated better registration accuracy, providing an objective measure of the alignment quality.

*4.7.2. Execution Time*

The computational efficiency of each algorithm was evaluated by measuring its execution time across the different dataset pairs. This metric was important for assessing the practicality of these methods for large-scale lunar mapping projects. Timing measurements included: Preprocessing time, Feature detection and description time, Feature matching time, Transformation estimation time and Image warping time. The registration accuracy and efficiency metrics demonstrated the effectiveness of our approach for creating integrated lunar datasets critical for scientific analysis and future mission planning.

## 5. RESULTS AND ANALYSIS

|  | OHRC - NAC | | IIRS - WAC | | DFSAR - SELENE | |
|---|---|---|---|---|---|---|
|  | Equatorial Region | Polar Region | Equatorial Region | Polar Region | Equatorial Region | Polar Region |
| AKAZE | ✓ | ✗ | ✓ | ✓ | ✗ | ✗ |
| ASIFT | ✓ | ✗ | ✓ | ✓ | ✗ | ✗ |
| RIFT2 | ✓ | ✗ | ✓ | ✓ | ✗ | ✗ |
| SIFT | ✓ | ✗ | ✓ | ✓ | ✗ | ✗ |
| SuperGlue | ✓ | ✓ | ✓ | ✓ | ✓ | ✓ |

*Table 2. Performance summary of algorithms in registering various dataset pairs*

Through enhanced preprocessing pipelines and rigorous evaluation using RMSE and execution time, the study confirms that SuperGlue delivers consistently superior accuracy and speed, particularly in equatorial and cross-modality cases, while traditional algorithms show regional and sensor-specific strengths but fail in extreme lunar conditions such as the poles. Table 2 presents the performance capability of all the algorithms in this research.

### 5.1. Qualitative Analysis

*5.1.1. AKAZE*

It consistently identifies a large number of keypoints and matches across all dataset pairs. This is particularly visible in the dense purple and blue match lines in subfigure (a) of figures 7, 8, and 9. Despite the high match count, several of these matches appear misaligned or loosely clustered around non-overlapping regions. This suggests a tendency to include outliers or less distinctive features, which could degrade final registration accuracy.

*5.1.2. RIFT2*

As seen in the yellow match lines in subfigure (d) of figures 7, 8, and 9, RIFT2 outputs fewer, but more precise correspondences. Matches appear to align accurately across structural features such as craters and ridges. It is particularly effective at handling the radiometric differences and structural dissimilarities between image pairs, suggesting strong performance in inter-sensor or inter-band alignment.

*5.1.3. SIFT & ASIFT*

Sparse but accurate (to an extent): Subfigures (b) and (c) in each dataset comparison reveal that SIFT and ASIFT detect relatively fewer matches. However, ASIFT slightly improves robustness to affine distortions over SIFT. The algorithms show limited effectiveness in complex terrains or when significant geometric distortion is present, especially in regions with shadow or occlusion.

*5.1.4. SuperGlue*

The final subfigures (e) in each dataset reveal that SuperGlue detects spatially coherent and geometrically accurate matches even in the presence of perspective distortions or occluded areas. Notably, it performs well even in challenging conditions with varying contrast or missing data (e.g., black stripes). It also performs well on OHRC-NAC polar datasets (Figure 10) and DFSAR-SELENE datasets (Figure 11), for which the other algorithms fail. Unlike classical descriptors, SuperGlue leverages global image context, allowing it to maintain correspondences across visually dissimilar yet structurally aligned regions.

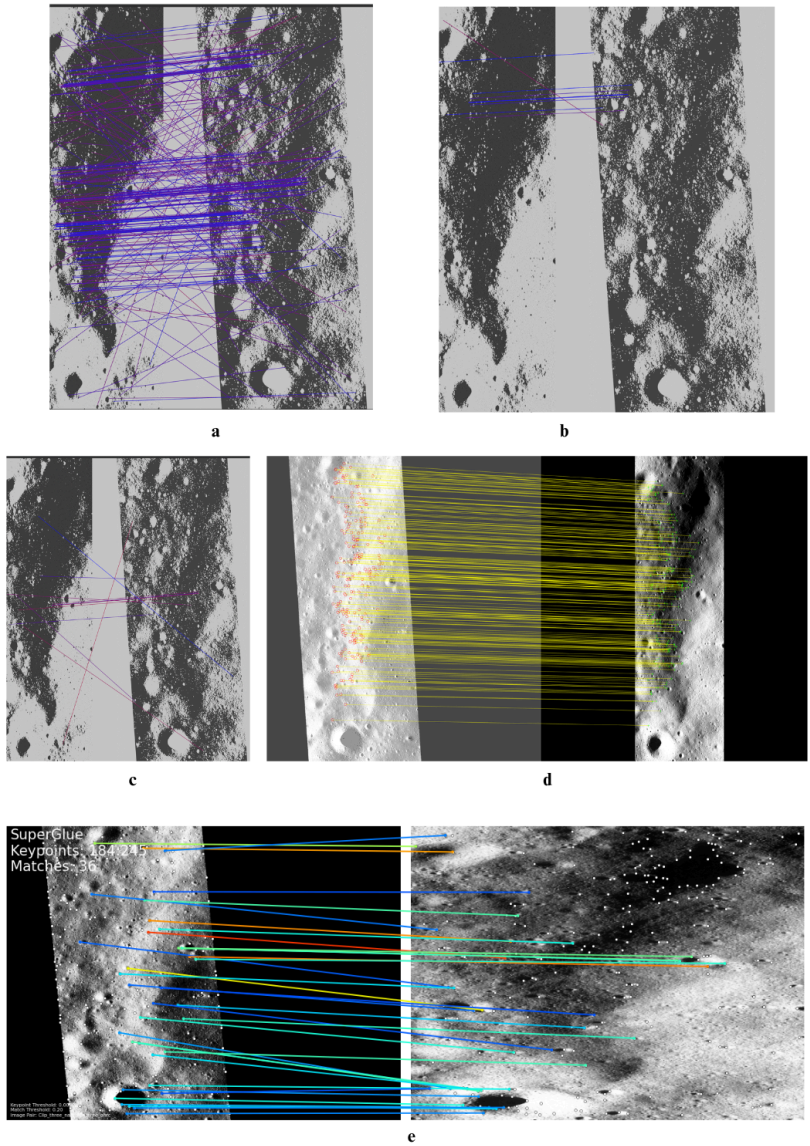

*Figure 7.* a) AKAZE b) ASIFT c) SIFT d) RIFT2 e) SuperGlue applied to OHRC-NAC Equatorial Dataset

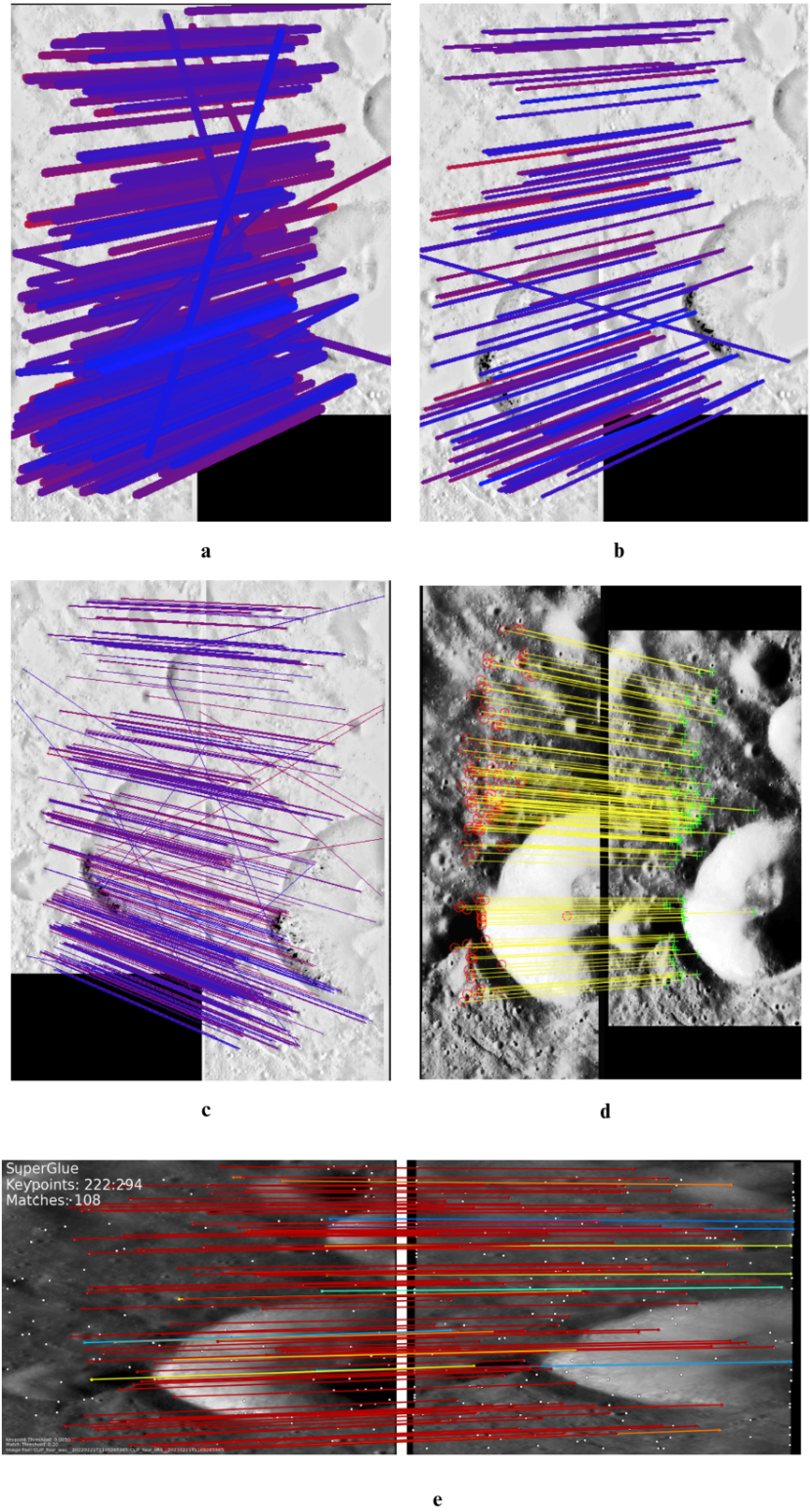

**Figure 8.** *a) AKAZE b) ASIFT c) SIFT d) RIFT2 e) SuperGlue applied to IIRS-WAC Equatorial Dataset*

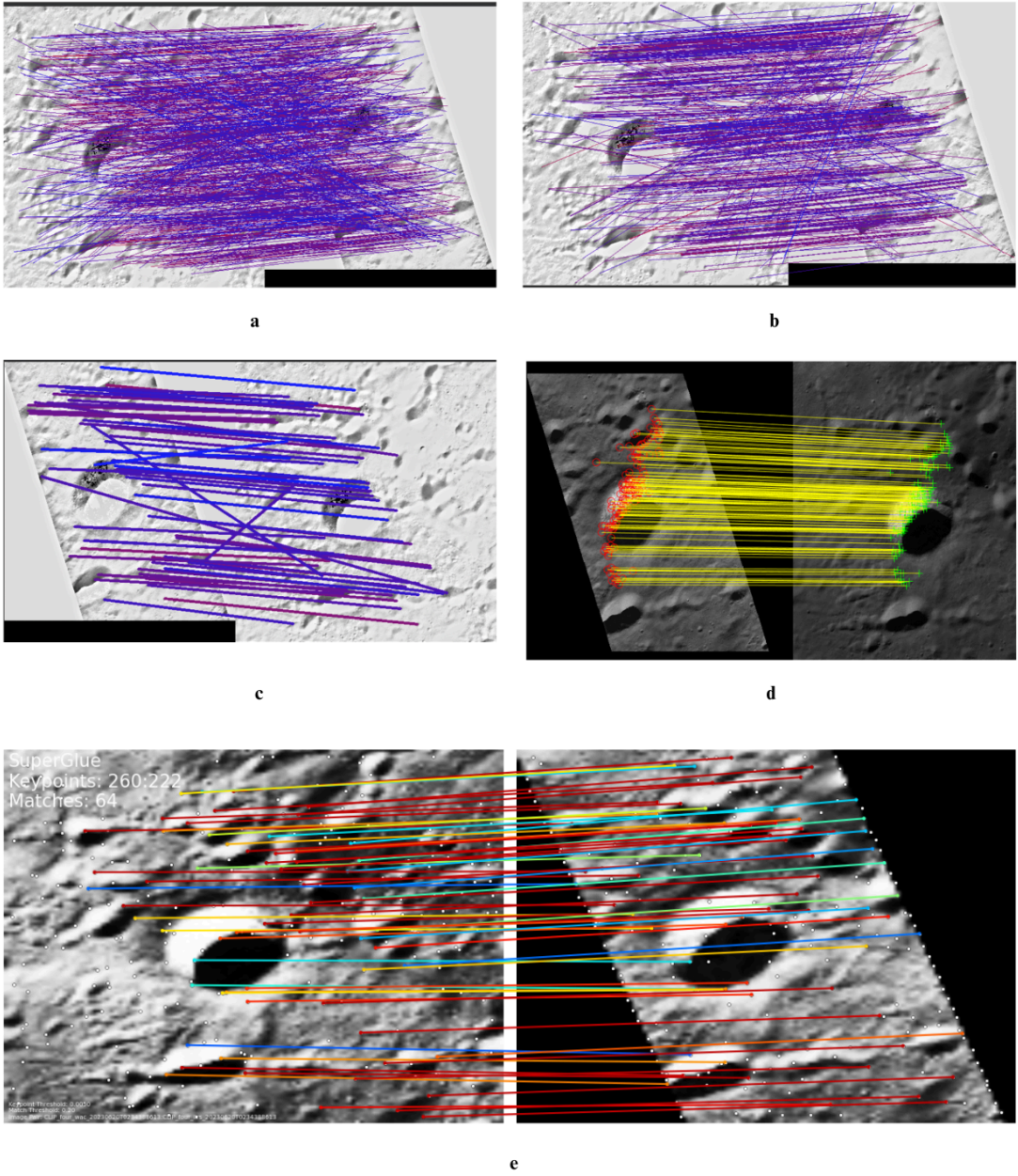

*Figure 9.* a) AKAZE b) ASIFT c) SIFT d) RIFT2 e) SuperGlue applied to IIRS-WAC Polar Dataset

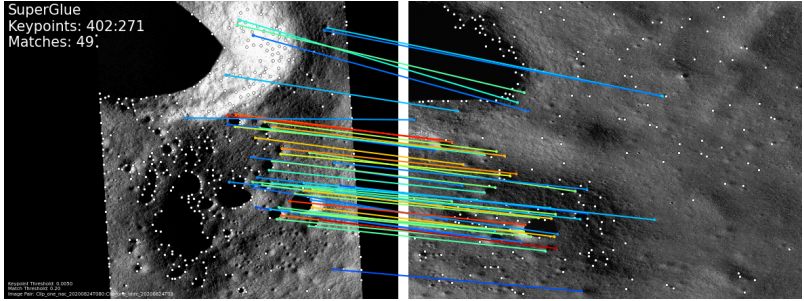
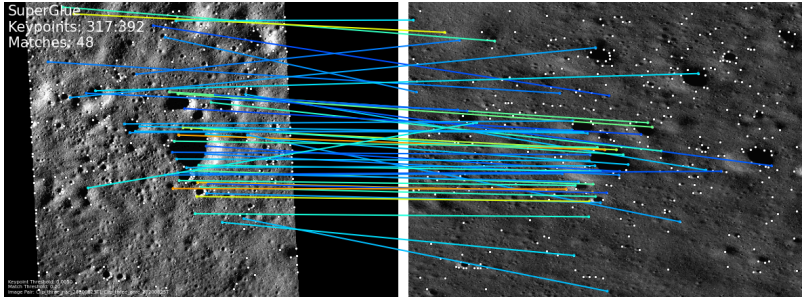

*Figure 10.* SuperGlue applied to OHRC-NAC Polar Datasets

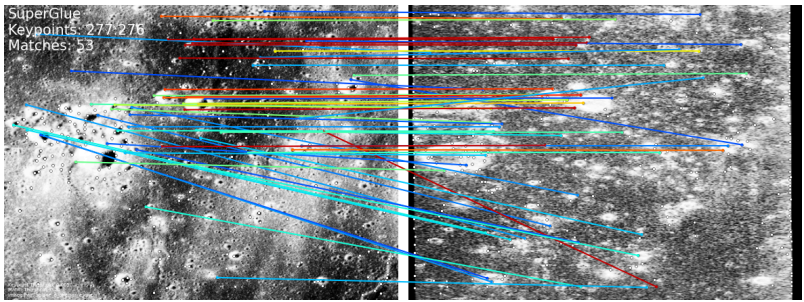
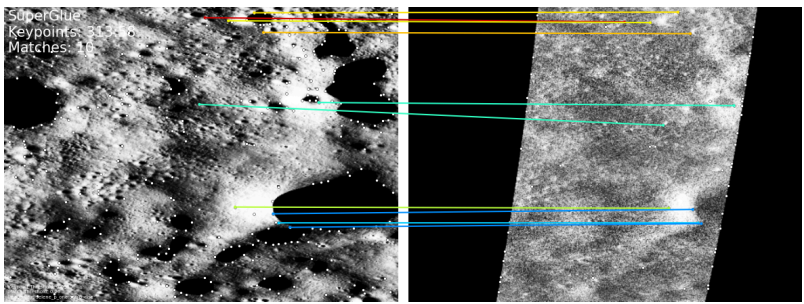

*Figure 11.* SuperGlue applied to DFSAR-SELENE Equatorial and Polar Datasets

## 5.2. Quantitative Analysis

Table 3 presents the Root Mean Square Error (RMSE) in the X and Y directions and the time taken for each of the five algorithms to perform image registration on the different lunar datasets:

|  | OHRC NAC Equatorial | | | OHRC NAC Polar | | |
|---|---|---|---|---|---|---|
| *ALGORITHMS* | *RMSE X* | *RMSE Y* | *TIME* | *RMSE X* | *RMSE Y* | *TIME* |
| **AKAZE** | 3.1189 | 4.7096 | 737.1722 | NA | NA | NA |
| **ASIFT** | 1.9946 | 1.6345 | 809.8209 | NA | NA | NA |
| **RIFT2** | 1.5033 | 1.1888 | 36.881 | NA | NA | NA |
| **SIFT** | 3.6096 | 5.9558 | 678.1992 | NA | NA | NA |
| **SuperGlue** | 0.6249 | 0.5718 | 3.809 | 0.9234 | 0.7586 | 4.643 |
|  | IIRS WAC Equatorial | | | IIRS WAC Polar | | |
| **AKAZE** | 0.5551 | 1.0841 | 0.1634 | 0.4595 | 1.3891 | 0.2287 |
| **ASIFT** | 0.5578 | 1.099 | 1.6407 | 0.9837 | 0.4501 | 2.0922 |
| **RIFT2** | 1.4056 | 1.3579 | 7.3894 | 0.7529 | 1.1054 | 12.716 |
| **SIFT** | 0.6879 | 1.1066 | 0.1207 | 2.0085 | 0.405 | 0.1607 |
| **SuperGlue** | 0.5069 | 0.6167 | 0.818 | 0.7681 | 0.9267 | 0.774 |
|  | DFSAR SELENE Equitorial | | | DFSAR SELENE Polar | | |
| **SuperGlue** | 0.3851 | 0.8432 | 0.745 | 0.9234 | 0.7586 | 4.643 |

***Table 3.*** *RMSE (in pixels) and Execution time (in seconds) of registered datasets.*

Analysing the RMSE values, which serve as an indicator of registration accuracy, reveals that for the OHRC NAC Equatorial dataset, SUPERGLUE achieves the lowest RMSE in both X and Y directions, signifying the highest accuracy among the tested algorithms. RIFT2 also demonstrates relatively good accuracy. In contrast, SIFT and AKAZE exhibit higher RMSE values, suggesting lower accuracy for this particular dataset. ASIFT also shows a relatively low RMSE, but still higher than SUPERGLUE and RIFT2. For the OHRC NAC Polar dataset, data is only available for SUPERGLUE, which shows RMSE X of 0.92347 and RMSE Y of 0.75861. The absence of data for the other algorithms on this polar dataset might indicate challenges encountered during the registration process, potentially due to geometric distortions.

On the IIRS WAC Equatorial dataset, SUPERGLUE again exhibits the lowest RMSE, closely followed by AKAZE and ASIFT. RIFT2 shows a slightly higher RMSE, mostly due to low intensity variations in the data, and SIFT has the highest RMSE among the algorithms with available data. For the IIRS WAC Polar dataset, SUPERGLUE maintains its lead with the lowest RMSE, followed by AKAZE. RIFT2 and SIFT show comparable yet higher RMSE values.

Regarding the SAR SELENE Equatorial and Polar datasets, SUPERGLUE achieves a low RMSE score. SIFT, ASIFT, RIFT2 and AKAZE failed to register these datasets.

Comparing the time taken for registration reveals significant differences in computational efficiency. For the OHRC NAC Equatorial dataset, SUPERGLUE is remarkably fast. RIFT2 is also relatively efficient, taking 36.881 seconds. In contrast, SIFT, AKAZE, and ASIFT require substantially longer processing times.

On the IIRS WAC Equatorial dataset, SUPERGLUE remains highly efficient. RIFT2 and AKAZE also exhibit relatively low processing times, while SIFT and ASIFT take longer. For the IIRS WAC Polar dataset, SUPERGLUE takes 0.774 seconds, and AKAZE is very fast at 0.2287 seconds. RIFT2 takes 12.716 seconds, and SIFT is also significantly fast at 0.1607 seconds.

For the SAR SELENE Equatorial dataset, SUPERGLUE takes 0.745 seconds, and for the Polar dataset, it takes 0.746 seconds.

Considering the algorithms that require preprocessing (SIFT, ASIFT, AKAZE) versus those that do not (RIFT2, SUPERGLUE), the data suggests that for the optical equatorial datasets (OHRC NAC and IIRS WAC), SUPERGLUE (no preprocessing) generally achieves superior accuracy and significantly better efficiency compared to the preprocessed algorithms. RIFT2 (no preprocessing) also demonstrates competitive performance in terms of accuracy and time. The high processing time associated with ASIFT for OHRC NAC Equatorial, despite its relatively good accuracy, highlights a notable trade-off. Overall, the data indicates that for these lunar datasets, algorithms that do not necessitate explicit preprocessing can indeed outperform traditional methods that do, both in terms of accuracy and computational efficiency.

# 6. CONCLUSION

Across the heterogeneous datasets registered, Superglue turned out to be the most efficient in terms of speed and accuracy. Its consistent results reflect upon the ability of deep learning algorithms to handle varied geometric distortions including the challenging OHRC-NAC polar dataset pair better than the traditional algorithms.

RIFT2 also showed potential for multimodal registration. However, it ran into trouble with some critical datasets, like OHRC–NAC Polar and SAR–SELENE Equatorial, revealing gaps in its robustness.

Traditional algorithms like SIFT and ASIFT, accurately registered less distorted data like the OHRC-NAC equatorial pair. More notably, both failed to deliver in the polar datasets, highlighting their sensitivity to the kinds of extreme distortions. AKAZE, while generally quicker than SIFT and ASIFT, also fell short on polar data. Even though the traditional algorithms registered certain datasets accurately, it was at a much higher computational cost than the deep learning algorithm.

What becomes clear is that many of the challenges in lunar image registration—especially in the polar regions—stem from significant viewpoint changes and variations in crater shapes, which traditional approaches aren't built to handle well. In contrast, SuperGlue's learning-based approach shows much stronger adaptability in these scenarios.

Based on these findings, the following recommendations can be made for selecting appropriate algorithms for lunar image registration:

- For applications requiring the highest accuracy and robustness across a variety of lunar datasets, including those with significant geometric distortions like the polar regions, SUPERGLUE appears to be the most suitable choice.
- RIFT2 can be considered for multimodal image registration tasks, particularly for the IIRS WAC Polar dataset, but its limitations with OHRC-NAC Polar and SAR SELENE data should be noted.
- SIFT, ASIFT, and AKAZE, while effective for optical equatorial imagery, are likely not the best options for polar regions with substantial angle changes and may have limitations with certain SAR datasets.
- Further investigation is needed to understand the specific conditions under which RIFT2 performs well and its limitations with different types of lunar imagery.

## 7. ACKNOWLEDGEMENT

The authors gratefully acknowledge the Space Applications Center, Indian Space Research Organisation (ISRO) for providing access to critical datasets and guidance that were instrumental in this research. We also extend our sincere thanks to Manipal University Jaipur for offering their support throughout the study. Their contributions significantly enhanced the quality and depth of our research. This collaboration was vital to the successful execution of the research work.

## 8. DECLARATION

To ensure linguistic clarity and academic tone, this manuscript underwent language refinement using Grammarly and SciSpace. Grammarly was employed to enhance grammar, punctuation, and overall readability, while SciSpace assisted in maintaining formal and scholarly writing standards. These tools contributed to improving the coherence and professionalism of the manuscript without altering its scientific content. All revisions were carefully reviewed to preserve the integrity and originality of the research.

## 9. REFERENCES

1. NASA. Image Registration for Remote Sensing. NASA Technical Reports Server. https://ntrs.nasa.gov/api/citations/20120008278/downloads/20120008278.pdf. Accessed 24 Apr 2025.
2. Paul S, Pati UC. A comprehensive review on remote sensing image registration. Int J Remote Sens. 2021;42:5400–5436.
3. Kumar, A., Kaushal, S., Murthy, S.V.: MoonMetaSync: Lunar Image Registration Analysis. In: 2024 IEEE Western New York Image and Signal Processing Workshop (WNYISPW), pp. 1–5 (2024).
4. Lowe, D.G.: Distinctive image features from scale-invariant keypoints. Int. J. Comput. Vis. 60, 91–110 (2004).
5. Ni, R., Zhao, F., Lu, P., Meng, T.: Spectral enhancement in lunar low-illumination regions: a CycleGAN-based approach for M3 data compensation. In: 55th Lunar and Planetary Science Conference (LPSC 2025), Abstract #1973. Lunar and Planetary Institute, Houston (2025).
6. Hurtik, P., Števuliáková, P., Perfilieva, I.: SIFT limitations in sub-image searching. Centre of Excellence IT4Innovations, Institute for Research and Applications of Fuzzy Modeling, University of Ostrava, Czech Republic (2017).
7. Wang, T., Wang, Y., Liang, Y., Huang, L., Yang, J., Li, W., & Ning, X. (2025). Feature point extraction for extra-affine image. arXiv.

8. Yu, G., & Morel, J.-M. (2011). ASIFT: An algorithm for fully affine invariant comparison. Image Processing On Line, 1.
9. Zhang, J., & Tang, B. (2024). Multi-temporal snow-covered remote sensing image matching via image transformation and multi-level feature extraction. Optics, 5(4), 392–405.
10. Li, J., Shi, P., Hu, Q., & Zhang, Y. (2023). RIFT2: Speeding-up RIFT with a new rotation-invariance technique. arXiv.
11. Soleimani, P., Capson, D., & Li, K. F. (2021). Real-time FPGA-based implementation of the AKAZE algorithm with nonlinear scale space generation using image partitioning. Journal of Real-Time Image Processing, 18.
12. Lozano-Vázquez, L., Miura, J., Rosales, A., Luviano-Juárez, A., & Mújica-Vargas, D. (2022). Analysis of different image enhancement and feature extraction methods. Mathematics, 10(14), 2407.
13. Tareen, S.K., & Saleem, Z. (2018). A comparative analysis of SIFT, SURF, KAZE, AKAZE, ORB, and BRISK. In Proceedings of the 2018 International Conference on Computing, Mathematics and Engineering Technologies (iCoMET).
14. Wang, A., Pruksachatkun, Y., Nangia, N., Singh, A., Michael, J., Hill, F., Levy, O., & Bowman, S.R. (2020). SuperGLUE: A stickier benchmark for general-purpose language understanding systems. *arXiv*.
15. Li, F., Chen, Y., Shi, Q., Shi, G., Yang, H., & Na, J. (2025). Improved low-light image feature matching algorithm based on the SuperGlue Net model. Remote Sensing, 17(5), 905.
16. Ge, H., Geng, Y., Ba, X., Wang, Y., & Lv, J. (2024). Automated registration of full moon remote sensing images based on triangulated network constraints. The International Archives of the Photogrammetry, Remote Sensing and Spatial Information Sciences, XLVIII-2-2024, 89–98.
17. Makharia, R., Singla, J.G., Amitabh, Dube, N., & Sharma, H. (2024). Image Registration of High Resolution Chandrayaan-2 Data. In Proceedings of the 2024 IEEE India Geoscience and Remote Sensing Symposium (InGARSS), XLVIII-2-2024, 1–4. IEEE.
18. Yang, Z., Kang, Z., & Yang, J. (2020). A semi-automatic registration method for Chang'E-1 IIM imagery based on globally geo-reference LROC-WAC mosaic imagery. IEEE Geoscience and Remote Sensing Letters, 18, 543–547.
19. Yang, Z., & Kang, Z. (2017). Accurate registration of the Chang'E-1 IIM data based on LRO LROC-WAC mosaic data. The International Archives of the Photogrammetry, Remote Sensing and Spatial Information Sciences, XLII-3/W1, 201–206.
20. He, B., Zheng, W., He, Z., & Ding, D. (2024). SAR simulation of lunar surface based on rocks abundance from NAC. In Proceedings of the PIERS 2024 Conference, April 2024, pp. 1–10.

21. Li, J., Hu, Q., & Ai, M. (2020). RIFT: Multi-modal image matching based on radiation-variation insensitive feature transform. IEEE Transactions on Image Processing, 29, 3296–3310.
22. Liu, X., Ding, Y. & Liu, C. (2025). MSIM: A multiscale iteration method for aerial image and satellite image registration. Remote Sensing, 17(8), 1423.
23. Chen, M., Tustison, N.J., Jena, R., et al. (2023). Image registration: Fundamentals and recent advances based on deep learning. In: Colliot, O. (ed.) Machine Learning for Brain Disorders, vol. 14. Humana, New York, NY.